\documentclass[10pt,twocolumn,letterpaper]{article}
\usepackage[T1]{fontenc}
\usepackage[pagenumbers]{cvpr}
\usepackage{cuted}
\usepackage{currfile}
\usepackage{caption}
\usepackage{flushend}
\raggedend
\usepackage{needspace}

\makeatletter
\iftoggle{cvprfinal}{}{%
    \apptocmd{\@maketitle}{\vspace*{29pt}}{}{%
        \PackageError{tess}{Could not preserve the manuscript title spacing}{}%
    }%
}
\makeatother

\usepackage{xspace}
\usepackage{array}
\usepackage{colortbl}
\usepackage{tabularx}
\usepackage{algorithm}
\usepackage{algpseudocode}
\usepackage{iftex}
\ifPDFTeX
\else
    \ifXeTeX
        \AtBeginDocument{}
    \fi
\fi
\definecolor{cvprblue}{rgb}{0.21,0.49,0.74}
\usepackage[pagebackref,breaklinks,colorlinks,allcolors=cvprblue]{hyperref}

\newcommand{\method}{TESS\xspace}
\newcommand{\normscore}{\mathcal{N}}
\newcommand{\rel}{\mathcal{R}_q}
\newcommand{\cond}{\mathcal{R}_r}
\newcommand{\novel}{\mathcal{D}}
\newcommand{\cover}{\mathcal{T}}
\definecolor{tessgray}{gray}{0.92}
\newcommand{\scoregain}[2]{\mbox{\textbf{#1}\textsuperscript{\ensuremath{\mathord{\uparrow}}\textbf{#2}}}}
\newcommand{\metricup}[1]{\mbox{\textbf{#1}\,\ensuremath{\uparrow}}}
\newcommand{\metricdown}[1]{\mbox{\textbf{#1}\,\ensuremath{\downarrow}}}
\newcolumntype{L}[1]{>{\raggedright\arraybackslash}p{#1}}
\newcolumntype{Y}{>{\centering\arraybackslash}X}
\title{TESS: Query-Guided Temporal Event-Structured Frame Selection\\for Long-Video Understanding}
\author{
    Qi Lu\textsuperscript{1,3},
    Shiyan Tong\textsuperscript{2,3},
    Shujie Li\textsuperscript{3},
    Sen Li\textsuperscript{3},
    Likun shi\textsuperscript{3},
    Maofei Que\textsuperscript{3},
    Xiaolong Jiang\textsuperscript{3},
    Yao Hu\textsuperscript{3}\\
    {\normalsize\textsuperscript{1}University of Science and Technology of China}\\
    {\normalsize\textsuperscript{2}Southeast University\quad\textsuperscript{3}Xiaohongshu}
}
\hypersetup{pdftitle={TESS: Query-Guided Temporal Event-Structured Frame Selection for Long-Video Understanding},pdfauthor={Qi Lu, Shiyan Tong, Shujie Li, Sen Li, Likun shi, Maofei Que, Xiaolong Jiang, Yao Hu},pdfsubject={Public preprint}}

\begin{document}
\maketitle
% Defer top floats from the opening column to the first-page right column.
\suppressfloats[t]
\begin{figure}[t]
    \centering
    \includegraphics[width=\columnwidth]{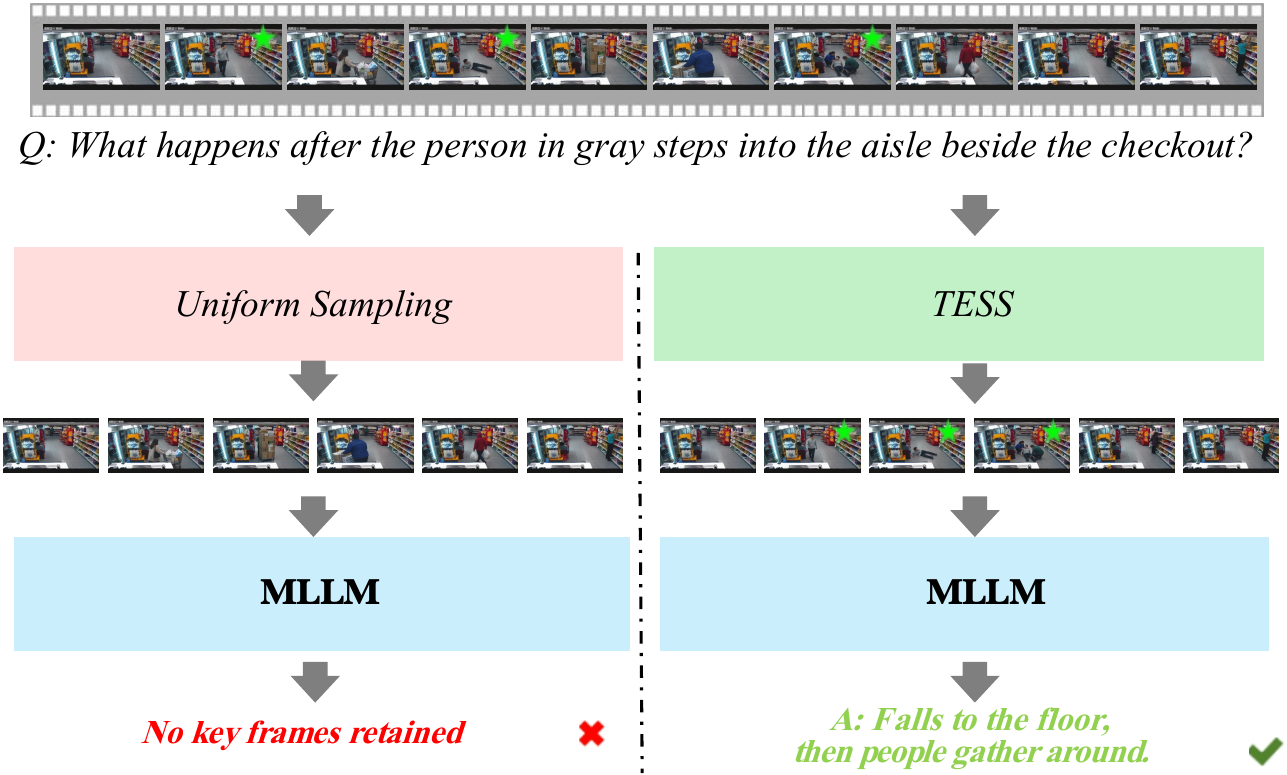}
    \caption{\textbf{Temporal evidence selection with \method.} In this illustrative comparison, uniform sampling misses the fall needed to answer what happens after the person enters the aisle. \method retains complementary observations of the anchor event and its aftermath in chronological order, supplying the evidence needed by the downstream model. Green stars mark query-relevant events.}
    \label{fig:contrast}
\end{figure}

\begin{abstract}
    Temporal localization in long videos is challenging because brief, question-relevant events are dispersed among extensive redundant content. Limited visual budgets increase the risk of omitting these events or their temporal context, undermining reliable reasoning. Existing frame selection methods emphasize semantic relevance or temporal coverage, often overlooking the complementary evidence needed to locate related events and establish their temporal relationships. To address this limitation, we propose TESS, a training-free framework for query-guided temporal event-structured frame selection. TESS selects frames that jointly show the reference event and the event asked about, in the required temporal order. Leveraging frozen vision-language representations, TESS integrates soft temporal gating, role-aware budget allocation, and within-role redundancy suppression to preserve complementary event evidence. Extensive experiments across four long-video understanding benchmarks and three multimodal backbones demonstrate that TESS consistently outperforms the strongest compared baseline by 4.0--5.5 percentage points under a fixed frame budget. Further analyses reveal greater improvements on longer videos and robust performance under tighter visual budgets, while independent evidence annotations confirm enhanced event and temporal-relation coverage. These findings establish explicit temporal evidence organization as an effective training-free strategy for preserving event relationships and supporting efficient temporal reasoning over long videos. Our code is available at \url{https://anonymous.4open.science/r/TESS_code}.
\end{abstract}

\section{Introduction}
\label{sec:introduction}

Localizing question-relevant events in long videos is fundamental to reasoning about their temporal relationships.
Recent multimodal large language models (MLLMs), including LLaVA-Video~\cite{zhang2025llavavideovideoinstructiontuning}, InternVL-3.5~\cite{wang2025internvl35advancingopensourcemultimodal}, and Qwen3-VL~\cite{bai2025qwen3vltechnicalreport}, have demonstrated strong capabilities in video understanding.
In videos spanning several minutes or longer, brief events are embedded in extensive redundant content, making their temporal localization difficult within limited visual budgets.
Benchmarks such as LongVideoBench~\cite{wu2024longvideobenchbenchmarklongcontextinterleaved} and Video-MME~\cite{fu2025videommefirstevercomprehensiveevaluation} highlight the difficulty of reasoning over events distributed across extended temporal contexts.
Effective frame selection must therefore localize related events and retain the surrounding observations needed to establish their order and transitions.

Existing frame selection methods address different aspects of efficient long-video understanding.
Adaptive Keyframe Sampling (AKS)~\cite{tang2025adaptivekeyframesamplinglong} balances query relevance and temporal coverage, while FOCUS~\cite{zhu2025focusefficientkeyframeselection} employs uncertainty-aware exploration to identify informative temporal regions.
Query- and Content-Aware selection (QCA)~\cite{peng2026qcaquerycontentawarekeyframe} allocates frames according to query relevance and content variation, followed by diversity-aware selection within temporal segments.
These approaches improve the relevance and distribution of selected frames, motivating further investigation of how the selected observations jointly support temporal reasoning.
In particular, understanding temporal relationships requires identifying relevant events and retaining complementary observations that establish their order, transitions, and contextual dependencies.
Coherent temporal evidence construction thus calls for coordinating semantic relevance, visual diversity, and temporal coverage around the relationships expressed in the query.

\Cref{fig:contrast} illustrates this challenge.
Understanding what happens after a person enters an aisle requires observations of both the entering event and the subsequent fall.
Retaining the fall alongside the person and surrounding scene provides the complementary evidence needed to establish the corresponding temporal relationship.
The roles of individual events depend on the query: the fall serves as the target event when examining the aftermath of entering and as the anchor event when reasoning about the subsequent response.
This observation motivates a frame selection strategy that organizes semantically relevant frames by query-dependent event roles, preserving the complementary observations needed to establish temporal relationships.

To address this challenge, we introduce \method, a training-free framework for query-guided temporal event-structured frame selection.
We formulate frame selection as relation-conditioned evidence set construction, decomposing each query into an anchor description, a target description, and their temporal relation.
Using frozen vision-language representations, soft temporal gating conditions target relevance on plausible anchor locations, while role-aware budget allocation preserves complementary anchor, target, transition, and contextual evidence.
Within-role redundancy suppression limits repeated evidence selection while preserving visually similar frames that support distinct temporal roles.
To improve evidence completeness, Core Evidence Repair retains observations associated with required temporal roles and supplements the selection with missing query-relevant, contextual, and temporal-pair evidence.
Finally, Constrained Set Refinement improves the selected evidence set while maintaining role and temporal coverage.
The resulting frames are presented in chronological order to a frozen downstream multimodal model, enabling direct integration into existing video question-answering pipelines.

We evaluate \method across four long-video understanding benchmarks and three multimodal backbones, using downstream question-answering accuracy to assess the effectiveness of selected visual evidence.
Under a fixed frame budget, \method achieves consistent improvements of 4.0--5.5 percentage points over the strongest compared baseline.
Further analyses investigate performance across video durations and frame budgets, computational efficiency, and individual component contributions.
Independent annotations of evidence intervals and temporal relations additionally assess the preservation of critical events and their temporal dependencies.
Our contributions are summarized as follows:
\begin{itemize}
    \item We formulate temporal frame selection as relation-conditioned evidence set construction, explicitly modeling query-dependent event roles and their temporal dependencies under a fixed visual budget.
    \item We propose \method, a training-free framework integrating soft temporal gating, role-aware evidence allocation, Core Evidence Repair, and Constrained Set Refinement to preserve complementary temporal evidence.
    \item The experimental results show that \method achieves superior performance across evaluation settings, surpassing the compared frame selection methods in long-video understanding and temporal reasoning.
\end{itemize}

\section{Related Work}
\label{sec:related-work}

\noindent\textbf{Video models and efficient representations.}
Flamingo~\cite{alayrac2022flamingo} supports interleaved visual--textual inputs through pretrained vision--language components, while BLIP-2~\cite{li2023blip2} bridges frozen image encoders and language models with a learned querying transformer. Visual instruction tuning in LLaVA~\cite{liu2023visualinstruction} further develops multimodal instruction following. Video-conditioned dialogue~\cite{li2025videochat,maaz2024videochatgpt}, audiovisual modeling~\cite{zhang2023videollama,cheng2024videollama2}, and aligned image--video representations~\cite{lin2024videollava} extend this paradigm to video, while LLaVA-Video~\cite{zhang2025llavavideovideoinstructiontuning} scales video instruction tuning with synthetic data. For longer sequences, MovieChat~\cite{song2024moviechat} consolidates dense video tokens into sparse memory, while MA-LMM~\cite{he2024malmm} maintains historical information in an online memory bank. LongVLM~\cite{weng2024longvlm} combines hierarchical token merging within segments with global context, and LongVU~\cite{shen2025longvu} uses frame redundancy, text guidance, and inter-frame dependencies for adaptive spatiotemporal compression. At the token level, Token Merging~\cite{bolya2023tome} combines similar visual tokens, and FastV~\cite{chen2024fastv} prunes them using early-layer attention. These approaches improve context utilization through representation learning, memory, or compression. Downstream reasoning draws on the visual evidence provided by the observations retained. \method addresses this input-selection stage by selecting RGB frames with frozen features and preserving the QA backbone's visual-input interface. Its relation-conditioned evidence objective complements representation-level compression by preserving complementary observations of temporally related events and their transitions.

% Queue the overview on page 2 for the top of page 3.
\begin{figure*}[t]
    \centering
    \includegraphics[width=\textwidth]{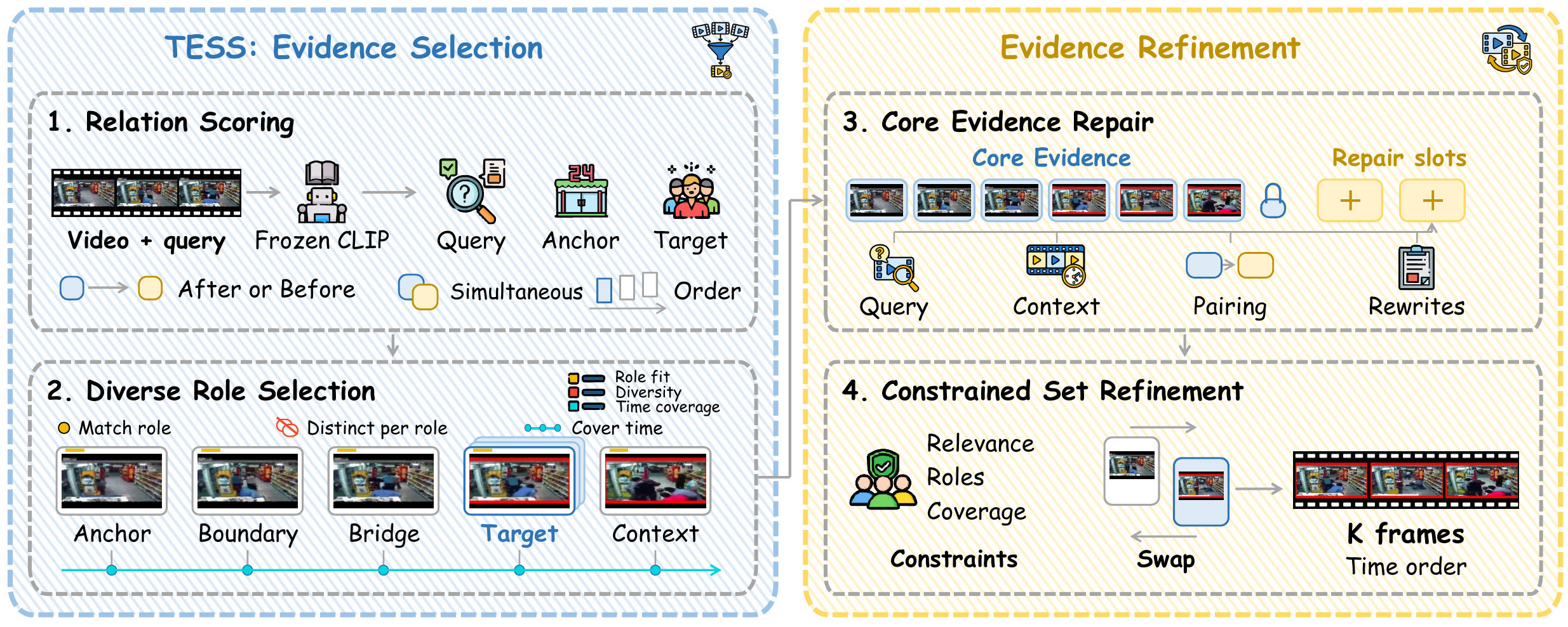}
    \caption{\textbf{Overview of \method.} Evidence selection (blue) uses frozen CLIP features for relation scoring and diverse selection across anchor, boundary, bridge, target, and context roles, balancing role fit, within-role diversity, and temporal coverage. Evidence refinement (yellow) includes Core Evidence Repair, which retains a role-aware core subset while supplementing query, context, and temporal-pair evidence. Query rewrites further augment relevance estimation. Constrained Set Refinement exchanges frames subject to relevance, role-coverage, and temporal-coverage constraints, returning $K$ frames in chronological order.}
    \label{fig:tess}
\end{figure*}

\noindent\textbf{Query-aware frame selection and retrieval.}
CLIP~\cite{pmlr-v139-radford21a}, BLIP~\cite{li2022blip}, and Long-CLIP~\cite{zhang2024longclip} provide representations or matching scores for query-conditioned retrieval. KeyVideoLLM~\cite{liang2024keyvideollm} selects keyframes through text--frame similarity, while Hu et al.~\cite{hu2025mllmselection} learn a selector using spatial supervision from a multimodal model and temporal supervision from a language model. Adaptive methods emphasize complementary selection criteria. AKS~\cite{tang2025adaptivekeyframesamplinglong} balances relevance and temporal coverage, and Q-Frame~\cite{zhang2025qframequeryawareframeselection} couples query-aware sampling with resolution adaptation. FOCUS~\cite{zhu2025focusefficientkeyframeselection} uses uncertainty-aware exploration to identify informative temporal clips before selecting frames within them, and QCA~\cite{peng2026qcaquerycontentawarekeyframe} allocates budgets by query relevance and content variation before diverse within-segment selection. Retrieval systems further use textual clip descriptions~\cite{ataallah2024goldfish}, query-adaptive hierarchies~\cite{wang2025videotree}, or iterative tool use~\cite{wang2024videoagent}. Temporal reasoning assigns distinct roles to visually similar observations, making their joint contribution to event relationships a key criterion for evidence selection. \method focuses on the composition of the evidence set: query-dependent roles distinguish anchors, targets, and contextual observations, while Core Evidence Repair and Constrained Set Refinement retain complementary observations needed to establish event relationships under a fixed frame budget.

\noindent\textbf{Temporal reasoning and evaluation.}
NExT-QA~\cite{xiao2021nextqa} evaluates causal and temporal action reasoning, while STAR~\cite{wu2021star} examines interaction, sequence, prediction, and feasibility. LongVideoBench~\cite{wu2024longvideobenchbenchmarklongcontextinterleaved} assesses reasoning within referred contexts, and Video-MME~\cite{fu2025videommefirstevercomprehensiveevaluation} spans diverse domains and durations. MLVU~\cite{zhou2025mlvubenchmarkingmultitasklong} covers multiple long-video tasks, while LVBench~\cite{wang2025lvbenchextremelongvideo} examines extended temporal contexts. These benchmarks connect event recognition with reasoning about relationships across time. For frame selection, answer accuracy also depends on which observations remain available to the model. We combine QA accuracy with annotated event and relation coverage to assess how selected observations support temporal reasoning.

% Finish Related Work at the column boundary; Method opens the next column.
\pagebreak[4]

\section{Method}
\label{sec:method}

We introduce \method, a query-guided framework for selecting temporally structured evidence from long videos. Answering a relational question requires complementary observations of the reference event, its transition, and the queried state to establish the requested temporal relationship. As shown in \cref{fig:tess}, \method comprises two stages: evidence selection combines relation scoring with diverse role selection, while evidence refinement performs Core Evidence Repair and Constrained Set Refinement. Both stages operate under a fixed frame budget, and the selected evidence retains its original temporal order.

\subsection{Problem Formulation and Preliminaries}
\label{sec:setup}

Given a video $V$ and a question $q$, we form the sequence $\mathcal{C}$ by uniformly sampling $N$ RGB frames $x_i$ with normalized timestamps $t_i$. With a frame budget $K$ bounded by the candidate count, we select
{\small
\begin{equation}
    S=\operatorname{TESS}(V,q;K)\subseteq\{1,\ldots,N\},\quad |S|=K,
    \label{eq:interface}
\end{equation}
}
such that the selected frames preserve complementary evidence for answering $q$. If the candidate sequence is shorter than the requested budget, all candidates are retained. The video and question define the selection inputs, enabling query-conditioned evidence construction as a dedicated stage preceding downstream answer generation.

A query-only parser extracts a temporal relation $r$, an anchor description $q_a$, a target description $q_g$, and an ordinal index when applicable. Its relation categories comprise directional relations (after and before), co-occurrence (simultaneous), and ordinal or sequence relations (first, last, nth, and order). The anchor defines the reference event, and the target describes the information sought by the question. Frozen CLIP encoders~\cite{pmlr-v139-radford21a} produce unit-normalized visual features $\mathbf v_i$ and text features $\mathbf e$ for each query variant, yielding
{\small
\begin{equation}
    s_i^z=\normscore\!\left(
        \{\mathbf v_j^\top\mathbf e(z)\}_{j=1}^{N}
    \right)_i,\quad z\in\{q,q_a,q_g\},
    \label{eq:similarity}
\end{equation}
}
where $\normscore$ denotes candidate-wise min--max normalization. We abbreviate these channels as $f_i$, $a_i$, and $g_i$, respectively, using $f_i$ as a fallback for missing anchor or target descriptions.

For long descriptive questions, query rewrites separately emphasize the task with its context and the event description. Their weighted responses replace $f_i$ during selection, repair, and refinement, broadening semantic matching while retaining the query's temporal relation.

\subsection{Relation-Conditioned Event Modeling}
\label{sec:relation}

\noindent\textbf{Directional evidence.}
For after and before relations, a softmax over anchor scores with temperature $\tau_a$ yields normalized anchor weights $p_a$. These weights aggregate soft temporal gates:
{\small
\begin{equation}
    G_i^r=\sum_j p_a(j)\,
    \sigma\!\left(\frac{\Delta_r(t_i,t_j)-\delta}{\tau_t}\right),
    \label{eq:temporal-gate}
\end{equation}
}
where $\Delta_r$ denotes target time minus anchor time for after queries and the reverse difference for before queries. The offset $\delta$ shifts the gate toward larger temporal separations, with $\tau_a$ and $\tau_t$ controlling weight concentration and gate softness. The gate scales target relevance to favor evidence on the appropriate side of plausible anchors.

Each candidate receives scores for five roles: anchor, boundary, bridge, target, and context. Anchor relevance identifies the reference event, with its local decline providing a boundary cue. Intermediate gate responses emphasize transitions for the bridge role, gated target relevance locates the queried event, and question relevance retains context. The normalized role response $R_i^\ell$ provides a soft localization cue for retrieving evidence assigned to role $\ell$.

\noindent\textbf{Simultaneous and ordinal evidence.}
For simultaneous events, the geometric mean of anchor and target responses favors joint evidence, and local averaging captures surrounding context. For ordinal and sequence queries, continuity scores are min--max-normalized maximum cosine similarities to the three frames ranked highest by summed anchor and query responses. We temporally merge and visually deduplicate their peaks to obtain ordered candidate occurrences. Target and context roles retain the requested occurrence and its surroundings, while early or late frames help establish first/last positions. These cues support co-occurrence and event-order reasoning. Queries lacking explicit relations use query relevance and contextual coverage.

% Keep role-aware selection at the next column after the shortened preliminaries.
\subsection{Role-Aware Evidence Selection}
\label{sec:roles}

Complementary event roles require explicit budget allocation to preserve brief transitions alongside dominant events. We assign quotas to the roles applicable to $r$, using role priors adjusted by their response strengths. Normalized allocation weights determine integer quotas $k_\ell$, with rounding residuals assigned in descending weight order so that the quotas sum to $K$.

Within each role, we greedily select unused candidates according to
{\small
\begin{equation}
    u_i^\ell(S)=\lambda_rR_i^\ell
        +\lambda_dD_i^\ell(S)+\lambda_cC_i(S),
    \label{eq:role-utility}
\end{equation}
}
where $D_i^\ell$ decreases with the largest visual similarity to already selected frames of the same role, and $C_i$ measures the temporal distance to the nearest selected frame. The nonnegative weights balance role relevance, visual novelty, and temporal coverage. Restricting strong redundancy penalties to the same role preserves similar-looking frames that support different temporal claims. Boundary neighborhoods provide additional candidates when capacity remains, and relation-conditioned relevance fills any remaining slots. The initial set $S_0$ contains $K$ unique frames.

% Queue the main comparison on page four so it appears atop page five.
\begin{table*}[t]
    \centering
    \caption{\textbf{Performance comparison with existing keyframe selection methods} on LongVideoBench, Video-MME, MLVU, and LVBench. We compare \method with uniform sampling, Top-$k$ selection of the most image--text matched frames, and recent frame selection baselines across LLaVA-Video-7B, InternVL-3.5-8B, and Qwen3-VL-8B, using a 64-frame budget. Scores are accuracies (\%), using M-Avg for MLVU. LLM size is in billions (B). $\dagger$ denotes our re-implementation. \textbf{Bold} and underlining indicate the best and second-best scores per backbone. Superscripts report gains over the strongest baseline in percentage points. Gray rows denote \method.}
    \label{tab:main-results}
    % The author confirmed that dagger-marked scores are our re-implementations.
    % Ranking formatting preserves all author-supplied values.
    \small
    \setlength{\tabcolsep}{3pt}
    \renewcommand{\arraystretch}{1.10}
    \begin{tabularx}{\textwidth}{@{}L{.27\textwidth}*{6}{Y}@{}}
        \toprule
        \textbf{Method} & \shortstack{\textbf{LLM}\\\textbf{size}} & \shortstack{\textbf{Scoring}\\\textbf{encoder}} & \shortstack{\textbf{LongVideo}\\\textbf{Bench}} & \shortstack{\textbf{Video}\\\textbf{MME}} & \shortstack{\textbf{MLVU}\\\textbf{M-Avg}} & \textbf{LVBench} \\
        \midrule
        \multicolumn{7}{l}{\textbf{LLaVA-Video-7B}} \\
        + Uniform & 7B & None & 58.9 & 64.4 & 70.8 & 41.9 \\
        + Top-$k$ & 7B & BLIP & 61.6 & 63.7 & 72.8 & 47.2 \\
        + AKS~\cite{tang2025adaptivekeyframesamplinglong} & 7B & BLIP & 62.7 & 65.3 & 71.8 & 47.6 \\
        + Q-Frame$^{\dagger}$~\cite{zhang2025qframequeryawareframeselection} & 7B & LongCLIP & 61.5 & 64.7 & 72.9 & 47.1 \\
        + FOCUS~\cite{zhu2025focusefficientkeyframeselection} & 7B & BLIP & \underline{63.5} & 65.4 & 72.7 & 48.2 \\
        + QCA~\cite{peng2026qcaquerycontentawarekeyframe} & 7B & BLIP & 62.9 & \underline{66.1} & \underline{74.1} & \underline{48.9} \\
        \rowcolor{tessgray}+ \textbf{TESS (Ours)} & \textbf{7B} & CLIP & \scoregain{67.5}{4.0} & \scoregain{70.5}{4.4} & \scoregain{78.9}{4.8} & \scoregain{54.3}{5.4} \\
        \midrule
        \multicolumn{7}{l}{\textbf{InternVL-3.5-8B}} \\
        + Uniform$^{\dagger}$ & 8B & None & 61.3 & 61.9 & 69.9 & 42.8 \\
        + Top-$k$ & 8B & BLIP & 62.5 & 61.6 & 70.4 & 48.7 \\
        + AKS$^{\dagger}$~\cite{tang2025adaptivekeyframesamplinglong} & 8B & BLIP & 62.9 & 62.8 & 70.5 & 47.9 \\
        + Q-Frame$^{\dagger}$~\cite{zhang2025qframequeryawareframeselection} & 8B & LongCLIP & 62.5 & 61.7 & 70.6 & 48.9 \\
        + FOCUS~\cite{zhu2025focusefficientkeyframeselection} & 8B & BLIP & \underline{63.8} & 63.1 & 70.8 & 49.4 \\
        + QCA~\cite{peng2026qcaquerycontentawarekeyframe} & 8B & BLIP & 63.5 & \underline{63.9} & \underline{71.3} & \underline{50.0} \\
        \rowcolor{tessgray}+ \textbf{TESS (Ours)} & \textbf{8B} & CLIP & \scoregain{68.2}{4.4} & \scoregain{68.1}{4.2} & \scoregain{76.2}{4.9} & \scoregain{55.2}{5.2} \\
        \midrule
        \multicolumn{7}{l}{\textbf{Qwen3-VL-8B}} \\
        + Uniform$^{\dagger}$ & 8B & None & 63.1 & 67.6 & 71.0 & 43.8 \\
        + Top-$k$ & 8B & BLIP & 64.4 & 67.4 & 74.2 & 50.7 \\
        + AKS$^{\dagger}$~\cite{tang2025adaptivekeyframesamplinglong} & 8B & BLIP & 64.7 & 68.6 & 74.2 & 50.8 \\
        + Q-Frame$^{\dagger}$~\cite{zhang2025qframequeryawareframeselection} & 8B & LongCLIP & 65.8 & 67.9 & 74.7 & 50.7 \\
        + FOCUS~\cite{zhu2025focusefficientkeyframeselection} & 8B & BLIP & 66.5 & 68.8 & 74.9 & 51.3 \\
        + QCA~\cite{peng2026qcaquerycontentawarekeyframe} & 8B & BLIP & \underline{66.9} & \underline{69.5} & \underline{75.7} & \underline{51.8} \\
        \rowcolor{tessgray}+ \textbf{TESS (Ours)} & \textbf{8B} & CLIP & \scoregain{71.8}{4.9} & \scoregain{73.8}{4.3} & \scoregain{80.8}{5.1} & \scoregain{57.3}{5.5} \\
        \bottomrule
    \end{tabularx}
\end{table*}

\subsection{Constrained Set Refinement}
\label{sec:refinement}

\noindent\textbf{Core Evidence Repair.}
To recover evidence omitted during initial selection, we retain a role-aware core subset $B$ and reserve the remaining budget for repair:
{\small
\begin{equation}
    |B|=\operatorname{round}(\rho K),\qquad K_R=K-|B|,
    \label{eq:repair-budget}
\end{equation}
}
where $\rho$ controls the retained fraction. High-scoring role representatives remain fixed while query and context repair replenish complementary observations. Relation-consistent pair repair then replaces mutable frames with compatible anchor--target evidence. Core retention and evidence recovery jointly preserve complementary temporal observations.

\noindent\textbf{Constrained Set Refinement.}
We optimize the repaired evidence set using
{\small
\begin{equation}
    J(S)=w_q\rel(S)+w_r\cond(S)+w_d\novel(S)+w_t\cover(S),
    \label{eq:set-objective}
\end{equation}
}
where nonnegative weights balance query relevance, role consistency, visual diversity, and temporal coverage. Each iteration selects the highest-gain feasible frame exchange subject to
{\small
\begin{equation}
    \Delta J>\eta,\quad
    \Delta\rel\geq-\epsilon_r,\quad
    \Delta\cover\geq0,
    \label{eq:exchange-constraints}
\end{equation}
}
where each $\Delta$ term denotes the change induced by the exchange, $\eta$ sets the minimum objective gain, and $\epsilon_r$ bounds the relevance decrease. Incoming frames inherit outgoing roles, and sole role representatives are protected. Core frames may be replaced during this stage when all constraints hold. The final $K$ unique frames are returned chronologically, using frozen features throughout.

\section{Experiments}
\label{sec:experiments}

Experiments across three QA backbones and four long-video benchmarks assess accuracy, duration sensitivity, frame efficiency, and computational cost, complemented by independent evidence diagnostics and component analyses.

\subsection{Experimental Setup}
\label{sec:experimental-setup}

\noindent\textbf{Datasets and Evaluation.}\@
The evaluation suite comprises LongVideoBench~\cite{wu2024longvideobenchbenchmarklongcontextinterleaved}, Video-MME~\cite{fu2025videommefirstevercomprehensiveevaluation}, MLVU~\cite{zhou2025mlvubenchmarkingmultitasklong}, and LVBench~\cite{wang2025lvbenchextremelongvideo}, covering referred reasoning, diverse video content, and extended temporal contexts. All methods follow official scoring protocols on identical benchmark splits and video--question manifests, with unparseable outputs marked incorrect. We adopt multiple-choice accuracy for LongVideoBench, Video-MME, and LVBench, and the multiple-choice task average (M-Avg) for MLVU.

These benchmarks test the generality of temporal evidence selection across tasks and video durations.

\noindent\textbf{Baselines.}\@
Comparisons include uniform sampling, query-similarity Top-$k$, and four adaptive selectors. AKS~\cite{tang2025adaptivekeyframesamplinglong} balances relevance and temporal coverage, and Q-Frame~\cite{zhang2025qframequeryawareframeselection} integrates query-aware selection with resolution adaptation. FOCUS~\cite{zhu2025focusefficientkeyframeselection} explores informative clips using uncertainty estimates, while QCA~\cite{peng2026qcaquerycontentawarekeyframe} allocates budgets by query relevance and content variation before diverse within-segment selection. Each method uses the scoring encoder listed in \cref{tab:main-results}, where our re-implementations are marked with $\dagger$.

\noindent\textbf{Implementation Details.}\@
Downstream QA uses LLaVA-Video-7B~\cite{zhang2025llavavideovideoinstructiontuning}, InternVL-3.5-8B~\cite{wang2025internvl35advancingopensourcemultimodal}, and Qwen3-VL-8B~\cite{bai2025qwen3vltechnicalreport} with a default budget of 64 frames. LLaVA-Video-7B supports the duration, budget, efficiency, and component analyses. \method uses frozen CLIP features to select query-relevant frames and supplies them chronologically for visual QA. For component ablations, candidate timestamps, preprocessing, QA prompts, decoding, and answer parsing are held fixed, together with selection hyperparameters and random seeds. Experiments run on eight NVIDIA H800 GPUs, with matched precision and batch size for efficiency comparisons.

\subsection{Performance Analysis}
\label{sec:performance-analysis}
\label{sec:subgroups}
\label{sec:frame-budget}
\label{sec:diagnostics-summary}

\noindent\textbf{Backbone and duration analysis.}\@
Table~\ref{tab:main-results} shows consistent improvements over uniform sampling across three backbones with 64 frames. With LLaVA-Video-7B, gains range from 6.1 to 12.4 percentage points across the four benchmarks. The LongVideoBench results in \Cref{tab:subgroup-results} further show that the advantage over the strongest baseline grows from a modest 1.1 points on short videos to 5.1 points on both medium and long videos. These results suggest that organizing complementary temporal evidence is particularly beneficial as video duration increases.

\begin{table}[t]
    \centering
    \caption{\textbf{Accuracy by video duration} on LongVideoBench with LLaVA-Video-7B and 64 frames. Short combines duration groups 15 and 60, while medium and long use groups 600 and 3600, respectively. Scores report accuracy (\%), with arrows indicating percentage-point gains over the best baseline per group.}
    \label{tab:subgroup-results}
    % Duration-group scores supplied by the user.
    \small
    \setlength{\tabcolsep}{2pt}
    \renewcommand{\arraystretch}{1.12}
    \begin{tabularx}{\columnwidth}{@{}L{.34\columnwidth}*{3}{Y}@{}}
        \toprule
        \textbf{Method} & \textbf{Short} & \textbf{Medium} & \textbf{Long} \\
        \midrule
        Uniform & 67.3 & 61.4 & 51.8 \\
        FOCUS & 71.2 & 65.5 & 57.1 \\
        QCA & 70.6 & 65.0 & 56.4 \\
        \rowcolor{tessgray}\textbf{TESS (Ours)} & \scoregain{72.3}{1.1} & \scoregain{70.6}{5.1} & \scoregain{62.2}{5.1} \\
        \bottomrule
    \end{tabularx}
\end{table}

\noindent\textbf{Frame-budget efficiency.}
\Cref{tab:frame-budget} shows that \method leads at every evaluated budget on both benchmarks with LLaVA-Video-7B. Using only 16 frames, it matches 64-frame FOCUS on LongVideoBench (63.5\%) and exceeds 64-frame QCA on Video-MME (66.9\% vs.\ 66.1\%). These results demonstrate that \method uses the available frame budget effectively, sustaining competitive QA accuracy with fewer selected observations.

\begin{table}[t]
    \centering
    \caption{\textbf{Frame-budget sensitivity with LLaVA-Video-7B} on LongVideoBench and Video-MME. Budget counts selected frames, with accuracy reported in percent. Arrows mark percentage-point gains over the best baseline at each budget. Candidate availability, resolution, and inference are fixed.}
    \label{tab:frame-budget}
    % Budget-sweep results supplied by the authors.
    \small
    \setlength{\tabcolsep}{2pt}
    \renewcommand{\arraystretch}{1.10}
    \begin{tabularx}{\columnwidth}{@{}L{.25\columnwidth}cYY@{}}
        \toprule
        \textbf{Method} & \textbf{Budget} & \textbf{LongVideoBench} & \textbf{Video-MME} \\
        \midrule
        Uniform & 16 & 53.6 & 59.8 \\
        FOCUS & 16 & 59.2 & 61.6 \\
        QCA & 16 & 58.8 & 62.6 \\
        \rowcolor{tessgray}\textbf{TESS (Ours)} & \textbf{16} & \scoregain{63.5}{4.3} & \scoregain{66.9}{4.3} \\
        \midrule
        Uniform & 32 & 56.4 & 62.3 \\
        FOCUS & 32 & 61.6 & 64.2 \\
        QCA & 32 & 61.3 & 64.7 \\
        \rowcolor{tessgray}\textbf{TESS (Ours)} & \textbf{32} & \scoregain{65.8}{4.2} & \scoregain{68.8}{4.1} \\
        \midrule
        Uniform & 64 & 58.9 & 64.4 \\
        FOCUS & 64 & 63.5 & 65.4 \\
        QCA & 64 & 62.9 & 66.1 \\
        \rowcolor{tessgray}\textbf{TESS (Ours)} & \textbf{64} & \scoregain{67.5}{4.0} & \scoregain{70.5}{4.4} \\
        \bottomrule
    \end{tabularx}
\end{table}

\subsection{Comparison with STATE-OF-THE-ART}
\label{sec:main-comparison}
\label{sec:efficiency}

\method ranks first across all 12 backbone--benchmark settings in \Cref{tab:main-results}, exceeding the strongest compared baseline by 4.0--5.5 percentage points. With Qwen3-VL-8B, it achieves 71.8\%, 73.8\%, 80.8\%, and 57.3\% on LongVideoBench, Video-MME, MLVU, and LVBench, respectively.

\noindent\textbf{Computational Efficiency.}\@
With LLaVA-Video-7B and 64 output frames, \Cref{tab:efficiency} shows that \method achieves the lowest end-to-end cost among the evaluated selectors. All methods use eight H800 GPUs with matched precision, batch size, and evaluation samples. Timing includes fresh feature computation, CPU stages, and auxiliary inference. \method reduces GPU hours by 34.6\% and 33.8\% relative to FOCUS on LongVideoBench and Video-MME, respectively. Under the candidate sampling configurations used in \Cref{tab:efficiency}, \method processes a smaller fraction of source frames than QCA. By preserving complementary temporal evidence while reducing overall computation, \method improves the accuracy--efficiency trade-off for long-video understanding.

\begin{table}[t]
    \centering
    \caption{\textbf{Efficiency with LLaVA-Video-7B and 64 frames.} E2E cost measures the complete inference pipeline in aggregate GPU hours on eight NVIDIA H800 GPUs. SFR is the percentage of source frames processed for feature extraction or scoring. Each method uses its own candidate sampling configuration; the output budget is fixed at 64 frames.}
    \label{tab:efficiency}
    % Efficiency measurements supplied by the user; current setup uses 8 H800 GPUs.
    \footnotesize
    \setlength{\tabcolsep}{2pt}
    \renewcommand{\arraystretch}{1.12}
    \begin{tabularx}{\columnwidth}{@{}L{.42\columnwidth}YY@{}}
        \toprule
        \textbf{Method} & \metricdown{SFR} & \metricdown{E2E cost} \\
        \midrule
        \multicolumn{3}{l}{\textbf{\textit{LongVideoBench}}} \\
        AKS & 3.7\% & 8.6 \\
        Q-Frame & 3.7\% & 6.4 \\
        FOCUS & 1.6\% & 5.2 \\
        QCA & 3.7\% & 7.8 \\
        \rowcolor{tessgray}\textbf{TESS (Ours)} & \textbf{1.5\%} & \textbf{3.4} \\
        \midrule
        \multicolumn{3}{l}{\textbf{\textit{Video-MME}}} \\
        AKS & 3.5\% & 11.2 \\
        Q-Frame & 3.5\% & 8.3 \\
        FOCUS & 1.6\% & 6.8 \\
        QCA & 3.5\% & 10.1 \\
        \rowcolor{tessgray}\textbf{TESS (Ours)} & \textbf{1.4\%} & \textbf{4.5} \\
        \bottomrule
    \end{tabularx}
\end{table}

% Queue the overview on page 6 so it accompanies Further Analysis on page 7.
\begin{figure*}[t]
    \centering
    \includegraphics[width=\textwidth]{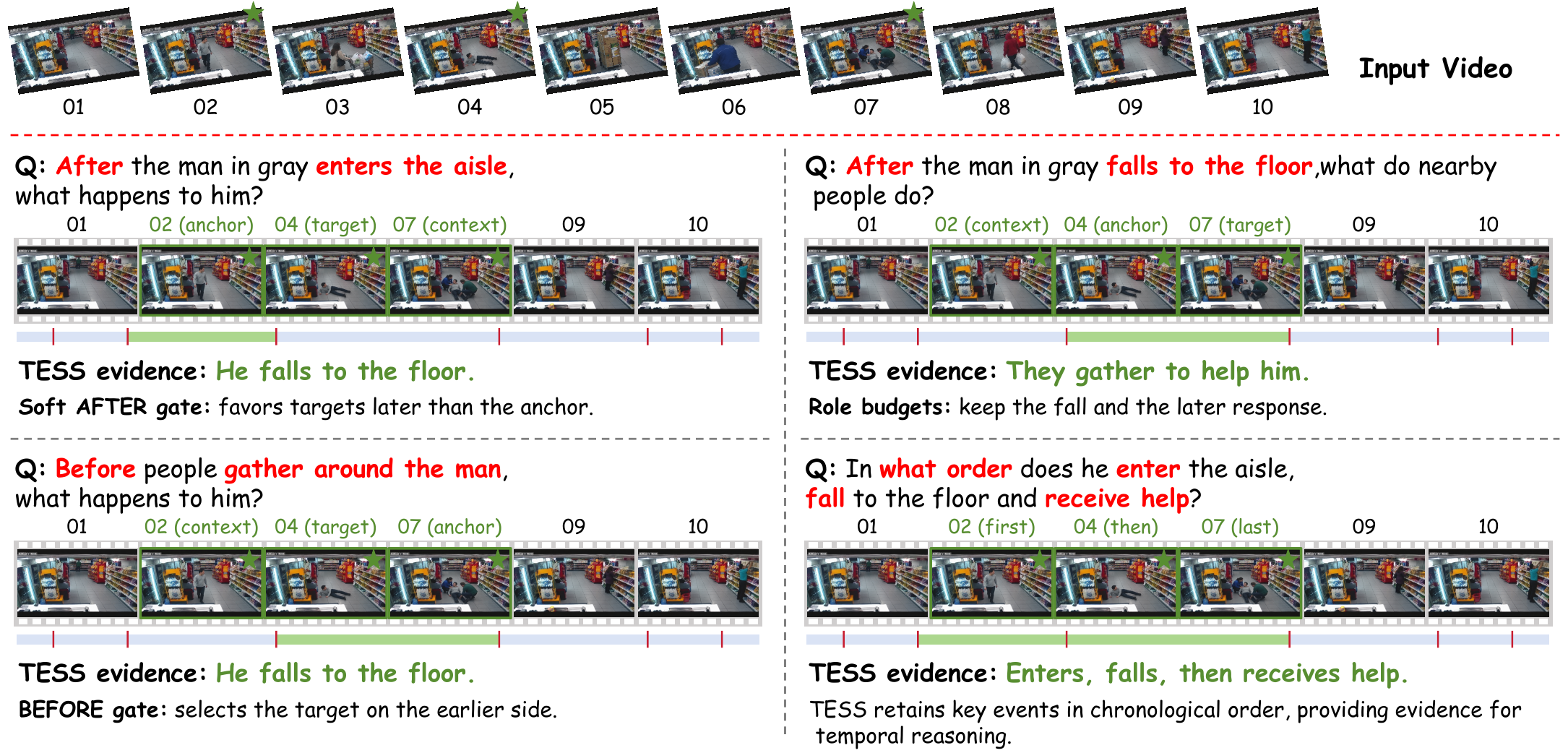}
    \caption{\textbf{Illustration of query-dependent temporal evidence.} The same video supports two after queries, a before query, and an event-ordering query. Each question specifies a reference event and its relation to the target. The fall serves as target evidence for the entering event and as anchor evidence for the subsequent response. Highlighted events therefore assume query-dependent roles, while selected frames retain the chronological sequence needed to interpret their relationships. Text below each strip describes the illustrated evidence.}
    \label{fig:temporal-analysis}
\end{figure*}

\subsection{Further Analysis}
\label{sec:further-analysis}

% Queue Table 5 for page 8 while Further Analysis flows continuously.
\begin{table*}[t]
    \centering
    \caption{\textbf{Component ablations of TESS} with LLaVA-Video-7B and 64 frames. Overall, Temporal, and Long report QA accuracy (\%) on all questions, independently annotated temporal questions, and long videos, respectively. Each ablation retains the remaining components and evaluation settings. The full method is shaded gray.}
    \label{tab:component-ablation}
    \footnotesize
    \setlength{\tabcolsep}{1.5pt}
    \renewcommand{\arraystretch}{1.18}
    \newcommand{\componentmetricwidth}{\dimexpr(\textwidth-.30\textwidth-12\tabcolsep-2\arrayrulewidth)/6\relax}
    \begin{tabular*}{\textwidth}{@{}L{.30\textwidth}|
        *{3}{>{\centering\arraybackslash}p{\componentmetricwidth}}|
        *{3}{>{\centering\arraybackslash}p{\componentmetricwidth}}@{}}
        \toprule
        \textbf{Variant} & \multicolumn{3}{c|}{\textbf{LongVideoBench}} & \multicolumn{3}{c}{\textbf{Video-MME}} \\
        \cmidrule(lr){2-4}\cmidrule(lr){5-7}
        & \metricup{Overall} & \metricup{Temporal} & \metricup{Long} & \metricup{Overall} & \metricup{Temporal} & \metricup{Long} \\
        \midrule
        w/o Temporal Gating & 64.6 & 61.5 & 58.6 & 67.8 & 63.6 & 60.2 \\
        w/o Role Allocation & 65.1 & 62.3 & 59.1 & 68.2 & 64.3 & 60.8 \\
        w/o Core Evidence Repair & 65.8 & 63.8 & 58.9 & 68.8 & 65.4 & 61.8 \\
        w/o Constrained Set Refinement & 66.4 & 64.7 & 60.3 & 69.5 & 66.2 & 62.3 \\
        \rowcolor{tessgray}\textbf{TESS (Full)} & \textbf{67.5} & \textbf{66.9} & \textbf{62.2} & \textbf{70.5} & \textbf{67.6} & \textbf{63.6} \\
        \bottomrule
    \end{tabular*}
\end{table*}

% Keep the temporal-evidence diagnostic on page 7; its discussion is in Ablation Studies.
% The author confirmed the currently plotted values as experimental results on 2026-10-19.
% Legacy asset filenames are retained to preserve existing links.
\begin{figure}[t]
    \centering
    \includegraphics[width=\columnwidth]{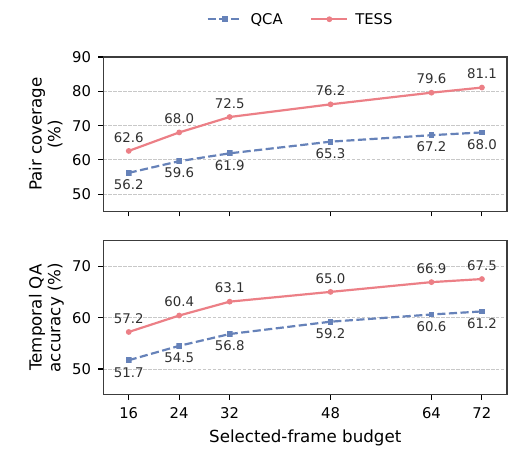}
    \caption{\textbf{Temporal evidence coverage and QA performance} on LongVideoBench with LLaVA-Video-7B. The upper and lower panels report relation-consistent pair coverage and temporal QA accuracy, respectively, across six frame budgets.}
    \label{fig:temporal-budget}
    \label{fig:temporal-budget-simulated}
    % Local spacing adjustment retains a clear caption-to-text separation.
    \vspace{-13pt}
\end{figure}

As illustrated in \cref{fig:temporal-analysis}, \method formulates temporal frame selection as relation-conditioned evidence set construction. The sequence of entering the aisle, falling, and receiving assistance supports distinct evidence roles across queries: the fall is target evidence when the anchor is entering the aisle and anchor evidence when the question concerns the subsequent response. A before query instead treats the gathering of nearby people as the anchor and the preceding fall as the target. Soft temporal gating prioritizes target evidence on the appropriate side of the anchor, making role assignment dependent on query semantics. Role-based budget allocation and within-role redundancy suppression retain complementary anchor, target, and contextual observations. Core Evidence Repair and Constrained Set Refinement preserve role and temporal coverage, while chronological output retains the event sequence required by ordering questions. These mechanisms provide the downstream model with semantically relevant and temporally organized evidence. By conditioning evidence roles on query semantics, we preserve the event relationships needed for temporal reasoning and provide the downstream model with a coherent visual basis for answering relational questions.

\section{Ablation Studies}
\label{sec:ablation-studies}

We ablate individual components to assess their contributions to temporal evidence construction and long-video reasoning. In \Cref{tab:component-ablation}, w/o Temporal Gating uses ungated target scores, and w/o Role Allocation removes role quotas while retaining role scores. w/o Core Evidence Repair disables query, context, and temporal-pair repair, and w/o Constrained Set Refinement omits exchanges after repair. TESS (Full) retains all components and serves as the reference.

All variants use LLaVA-Video-7B with frozen CLIP features and a 64-frame budget on LongVideoBench and Video-MME. All component ablations retain the same candidate pool and downstream inference settings as the full method. Fixed selection hyperparameters enable direct comparisons of component effects. We measure overall QA accuracy alongside temporal-question and long-video accuracy to assess reasoning over event relations and extended contexts. The complementary budget analysis in \Cref{fig:temporal-budget} measures relation-consistent pair coverage on independently annotated LongVideoBench questions. A question is covered when selected frames represent both reference events and satisfy their annotated temporal relation. This diagnostic directly assesses the evidence retained for relational reasoning.

The full method achieves the highest accuracy in every category in \Cref{tab:component-ablation}. Temporal gating and role allocation yield the largest gains in temporal QA accuracy. Removing gating reduces accuracy by 5.4 points on LongVideoBench and 4.0 points on Video-MME, while removing role allocation incurs losses of 4.6 and 3.3 points, respectively. These reductions underscore the importance of relation-conditioned localization and explicit role budgets for temporal reasoning. Repair and refinement provide further benefits after initial selection. Removing repair lowers long-video accuracy on LongVideoBench by 3.3 points, highlighting the contribution of recovered evidence. Removing refinement lowers this accuracy by 1.9 points, demonstrating the value of improving the selected set after repair.

\noindent\textbf{Evidence coverage across frame budgets.}
\Cref{fig:temporal-budget} complements the component ablations by examining evidence coverage and temporal QA across six frame budgets. \method consistently outperforms QCA in both metrics. The pair-coverage gain increases from 6.4 percentage points at 16 frames to 13.1 points at 72 frames, accompanied by temporal-QA gains of 5.5--6.3 points. At 64 frames, \method achieves 79.6\% pair coverage and 66.9\% temporal QA accuracy, compared with 67.2\% and 60.6\% for QCA. The advantage at 16 frames demonstrates effective evidence selection under a tight visual budget. As the budget increases, the larger coverage margin indicates that additional observations retain complementary events, extending the temporal evidence available to the question-answering model. This pattern supports allocating added capacity according to event roles and relationships. Pair coverage measures the retention of relation-consistent events, while temporal QA assesses the quality of answers derived from the selected observations. Their concurrent improvements connect more complete event evidence with stronger temporal reasoning under matched frame budgets. The controlled accuracy reductions in \Cref{tab:component-ablation} further establish the contribution of each core component to the complete selection pipeline. These results support a staged design in which localization and role allocation identify complementary events, repair restores missing observations, and refinement improves the composition of the selected evidence to support temporal reasoning under a fixed frame budget.

\section{Conclusion}
\label{sec:conclusion}

In this paper, we presented \method, a training-free framework for locating related events in long videos under limited visual budgets. By allocating frames to complementary event roles and recovering omitted observations, we preserve brief events and their temporal context amid extensive visual redundancy. Experiments across four benchmarks and three QA backbones demonstrate consistent improvements, with stronger gains on longer videos. Independent evidence measurements and controlled ablations further connect these gains to more complete temporal support, validating the importance of both evidence selection and subsequent refinement. Across frame budgets, we retain complementary observations that support temporal QA under constrained visual input. Frozen visual representations and chronologically ordered frame sequences facilitate integration with existing video question-answering models, preserving their visual input interface and supporting reuse across backbones. We conclude that structured evidence selection supports efficient long-video reasoning by preserving complementary observations and event relationships, providing semantically relevant and temporally coherent visual evidence under a limited frame budget.

\clearpage
% Reset column height after any full-width floats flushed from the main text.
\twocolumn
{
    \small
    \bibliographystyle{ieeenat_fullname}
    \bibliography{references}
}
\clearpage
\appendix
% Supplementary definitions have their own equation numbers.
\setcounter{equation}{0}
\renewcommand{\theequation}{S\arabic{equation}}
\renewcommand{\theHequation}{supp.\arabic{equation}}
\setcounter{table}{0}
\renewcommand{\thetable}{S\arabic{table}}
\renewcommand{\theHtable}{supp.\arabic{table}}
\setcounter{figure}{0}
\renewcommand{\thefigure}{S\arabic{figure}}
\renewcommand{\theHfigure}{supp.\arabic{figure}}
% Preserve native CVPR float spacing in the supplementary material.
\twocolumn
% Align complete columns using the native CVPR spacing flexibility.
\flushbottom
\setlength{\parskip}{0pt plus 1pt}
\section{More Related Work}
\label{sec:more-related-work}
Language-conditioned temporal grounding studies how textual descriptions identify events within continuous video. CTRL~\cite{gao2017tall} combines cross-modal matching with temporal boundary regression, while Moment Context Network~\cite{hendricks2017localizing} integrates local and global visual context to retrieve described moments. Moment-DETR~\cite{lei2021momentdetr} jointly predicts relevant temporal intervals and clip saliency through set prediction. TimeChat~\cite{ren2024timechat} introduces timestamp-aware frame encoding and a sliding video Q-Former for temporally localized video-language tasks. These approaches emphasize correspondence between language and event intervals. \method addresses the complementary question of which observations from those events should occupy a limited visual budget, organizing anchor, target, and contextual evidence for downstream reasoning. This connection motivates evaluating both event coverage and the temporal relationships preserved by a selected frame set.

\section{Details of Core Evidence Repair and Constrained Set Refinement}
\label{sec:repair-algorithm}
Evidence refinement combines recovery of missing temporal observations with constrained optimization of the selected frame set. \Cref{alg:core-repair} operates on frozen candidate features, timestamps, query and role scores, and the initial selection. The limits $M_p$ and $M_s$ bound pair-repair proposals and refinement iterations, respectively, and $\gamma_p$ specifies the minimum score improvement required for pair repair.
\par
\begin{algorithm}[t]
    \caption{Core Evidence Repair and Constrained Set Refinement}
    \label{alg:core-repair}
    \normalsize
    \algrenewcommand\algorithmicindent{1em}
    \algrenewcommand\algorithmicrequire{\textbf{Input:}}
    \algrenewcommand\algorithmicensure{\textbf{Output:}}
    \begin{algorithmic}
        \Require Candidates $\mathcal C$, features, scores, timestamps; initial set $S_0$ and roles; relation $r$, target $q_g$; budget $K$; transition-repair flag
        \Ensure $K$ distinct frames in temporal order
        \Statex \textbf{Core Evidence Repair}
        \State Retain role-aware core $B\subseteq S_0$ using Eq.~\eqref{eq:repair-budget} and initialize $S\gets B$
        \State Fill $K_R$ query/context slots by relevance, novelty, and temporal coverage
        \State Initialize mutable indices $\mathcal U\gets S-B$
        \State Extract up to $M_p$ relation-consistent proposals from smoothed anchor and target responses
        \For{each unused proposal $j$ while $\mathcal U\ne\varnothing$}
            \State $i\gets$ lowest-scoring frame in $\mathcal U$
            \If{candidate-score gain $\geq\gamma_p$}
                \State $S\gets(S-\{i\})\cup\{j\}$, $\mathcal U\gets\mathcal U-\{i\}$
                \State Assign the proposal role to $j$
            \EndIf
        \EndFor
        \Statex \textbf{Constrained Set Refinement}
        \State Build a ranked candidate pool including video endpoints
        \For{$m=1,\ldots,M_s$}
            \State Form feasible sets $\mathcal E(S)$ by single-frame exchanges with role inheritance, sole-role protection, and Eq.~\eqref{eq:exchange-constraints}
            \State \textbf{if} $\mathcal E(S)=\varnothing$ \textbf{then break}
            \State $S\gets\operatorname*{arg\,max}_{S'\in\mathcal E(S)} J(S')$ using Eq.~\eqref{eq:set-objective}
        \EndFor
        \If{transition repair enabled, $r=\textit{after}$, $q_g$ generic, and $B\subseteq S$}
            \State Repair transitions in $S$, keeping $B$ fixed
        \EndIf
        \State \Return $S$ sorted by timestamp
    \end{algorithmic}
\end{algorithm}

\noindent\textbf{Role-aware selection scores.}\@
Role-aware selection balances semantic relevance, within-role visual novelty, and temporal spacing. Let $\ell_i$ denote the role assigned to a selected frame. Unit-normalized visual features define cosine similarity, and selected frames are grouped by role:
{\small
\begin{equation}
    c_{ij}=\mathbf v_i^\top\mathbf v_j,\qquad
    S_\ell=\{j\in S:\ell_j=\ell\}.
    \label{eq:supp-role-set}
\end{equation}
}
For a nonempty selection, novelty and temporal coverage are
{\small
\begin{equation}
    \begin{aligned}
        D_i^\ell(S)&=1-\beta\max_{j\in S_\ell}[c_{ij}]_+,\\
        C_i(S)&=\min_{j\in S}|t_i-t_j|.
    \end{aligned}
    \label{eq:supp-selection-scores}
\end{equation}
}
Here, the positive-part operator replaces negative similarities with zero. The nonnegative coefficient $\beta$ controls within-role redundancy suppression, with larger values penalizing visually similar candidates more strongly. Restricting this penalty to $S_\ell$ allows similar observations to support distinct event roles. The coverage term $C_i$ uses all selected timestamps, normalized to the unit interval, to favor uncovered temporal regions. The weights $\lambda_r$, $\lambda_d$, and $\lambda_c$ balance role relevance, novelty, and temporal spacing. The effective similarity penalty scales with the product of $\lambda_d$ and $\beta$.

Selection starts from an empty set and processes roles in descending quota order. The maximum over an empty role set is defined as zero, initializing $D_i^\ell$ to one. Before any frame is selected, $C_i$ is also one, so the first choice is determined by role relevance. Subsequent choices favor complementary observations through their visual and temporal distances from the evolving selection, with ties resolved by candidate order.

\noindent\textbf{Core retention and budget allocation.}\@
We retain high-scoring representatives of required roles and preferentially remove locally redundant observations. Temporal weighting reduces the redundancy penalty for visually similar events separated in time. Query repair prioritizes relevance and visual novelty, and context repair emphasizes temporal spacing and boundary evidence. Relation-specific quotas allocate additional context to ordering queries. Reused frames retain their roles, and newly selected frames receive the highest-scoring available role.
\par

% Queue Table S1 before the continuous discussion of selection quality.
\begin{table*}[t]
    \centering
    \caption{\textbf{Independent evaluation of selected evidence} on LongVideoBench at a 64-frame budget. Evidence recall, complete coverage, and relation-consistent pair coverage are percentages. Redundancy and duration-normalized maximum gap are dimensionless. The upper rows give eligible question counts and candidate-pool coverage ceilings for each metric.}
    \label{tab:selection-diagnostics}
    \small
    \setlength{\tabcolsep}{3pt}
    \renewcommand{\arraystretch}{1.15}
    \begin{tabularx}{\textwidth}{@{}L{.26\textwidth}*{5}{Y}@{}}
        \toprule
        \textbf{Method} & \shortstack{\textbf{Evidence}\\\metricup{recall}} & \shortstack{\textbf{Complete}\\\metricup{coverage}} & \shortstack{\textbf{Pair}\\\metricup{coverage}} & \metricdown{Redundancy} & \shortstack{\textbf{Maximum}\\\metricdown{gap}} \\
        \midrule
        \multicolumn{6}{l}{\textbf{\textit{Annotations and candidate pool}}} \\
        Eligible questions & 420 & 420 & 265 & 420 & 420 \\
        Candidate-pool recall & 96.4 & 93.8 & 92.1 & -- & -- \\
        \midrule
        Uniform & 61.2 & 46.4 & 41.9 & 0.71 & 0.016 \\
        Top-$k$ & 68.4 & 52.1 & 47.5 & 0.86 & 0.142 \\
        AKS & 71.5 & 56.9 & 52.5 & 0.78 & 0.084 \\
        Q-Frame & 70.8 & 56.0 & 51.3 & 0.79 & 0.089 \\
        FOCUS & 72.7 & 58.1 & 54.0 & 0.75 & 0.068 \\
        QCA & 73.6 & 59.5 & 67.2 & 0.74 & 0.062 \\
        \rowcolor{tessgray}\textbf{TESS (Ours)} & \textbf{79.1} & \textbf{65.0} & \textbf{79.6} & \textbf{0.68} & \textbf{0.045} \\
        \bottomrule
    \end{tabularx}
\end{table*}

\noindent\textbf{Temporal-pair repair.}\@
Smoothed anchor and target responses identify compatible event pairs. Directional pairs satisfy the requested order and minimum lag, and simultaneous pairs favor temporal overlap. The selected pair provides frames at score peaks and in their neighborhoods, with first/last queries additionally favoring the corresponding temporal boundary. We accept an unused candidate when it improves on a mutable observation by at least $\gamma_p$, protecting the retained core throughout repair. Pair-repair candidates require both an applicable temporal relation and an available compatible pair.

\noindent\textbf{Constrained Set Refinement.}\@
For a nonempty selected set, query relevance and role consistency are defined as the mean normalized query and assigned-role responses:
{\small
\begin{equation}
    \rel(S)=\frac{1}{|S|}\sum_{i\in S}f_i,\qquad
    \cond(S)=\frac{1}{|S|}\sum_{i\in S}R_i^{\ell_i}.
    \label{eq:supp-set-relevance}
\end{equation}
}
Target-conditioned responses supply fallback scores for roles lacking dedicated scoring functions. These averages preserve query alignment and the semantic contribution of assigned evidence roles during refinement.

For sets containing at least two frames, global visual diversity is
{\small
\begin{equation}
    \novel(S)=\frac{1}{|S|}\sum_{i\in S}
    \left[1-\operatorname{clip}_{[0,1]}\!\left(
        \max_{\substack{j\in S\\j\ne i}}c_{ij}\right)\right].
    \label{eq:supp-set-diversity}
\end{equation}
}
Each frame is compared with its most similar other frame across all roles. Clipping cosine similarity to the unit interval bounds each novelty contribution and prevents negative similarities from yielding additional reward. Averaging these contributions penalizes redundant observations throughout the set. A singleton receives diversity one because it contains no pairwise repetition.

Let $m$ be the number of selected frames and $\widetilde t_k$ their timestamps in ascending order. Temporal coverage includes both video boundaries:
{\small
\begin{equation}
    \begin{aligned}
        \widetilde t_0&=0,\qquad \widetilde t_{m+1}=1,\\
        \cover(S)&=1-\max_{0\leq k\leq m}
            (\widetilde t_{k+1}-\widetilde t_k).
    \end{aligned}
    \label{eq:supp-set-coverage}
\end{equation}
}
Maximizing this term reduces the longest uncovered interval, complementing the local spacing criterion used during greedy selection. For an empty set, all four set metrics are defined as zero.

The nonnegative weights $w_q$, $w_r$, $w_d$, and $w_t$ balance query alignment, role consistency, global diversity, and temporal coverage, respectively. A bounded candidate pool supplies exchanges ranked by query and role relevance. Incoming frames inherit outgoing roles, preserving role counts, while the acceptance constraints bound relevance loss and prevent increased temporal gaps. Core frames remain eligible for replacement under these constraints, enabling improvements in global diversity while maintaining the required evidence roles.

\noindent\textbf{Optional transition repair.}\@
When enabled, transition repair applies to generic-target after queries only if the refined set retains the entire original core. Otherwise, the refined set is returned unchanged. Adjacent-feature changes provide approximate transition cues for replacing non-core frames while preserving occupied temporal bins and satisfying relevance and maximum-gap constraints.
\par

\par
\section{Independent Selection Quality}
\label{sec:supplementary-experiments}
\label{sec:selection-diagnostics}
% Queue the component table for the following page beside its discussion.
\begin{table*}[t]
    \centering
    \caption{\textbf{Component diagnostics and shared-encoder comparisons} on LongVideoBench with LLaVA-Video-7B and 64 frames. QA accuracy and coverage metrics are percentages. Sel. time denotes seconds per query, including encoding and query processing. Evidence metrics use the diagnostic subset in \cref{sec:selection-diagnostics}, and the lower panel uses shared frozen CLIP features.}
    \label{tab:ablation}
    \small
    \setlength{\tabcolsep}{2.5pt}
    \renewcommand{\arraystretch}{1.15}
    \begin{tabularx}{\textwidth}{@{}L{.30\textwidth}*{6}{Y}@{}}
        \toprule
        \textbf{Variant} & \shortstack{\textbf{Overall}\\\metricup{QA}} & \shortstack{\textbf{Temporal}\\\metricup{QA}} & \shortstack{\textbf{Long}\\\metricup{QA}} & \shortstack{\textbf{Evidence}\\\metricup{recall}} & \shortstack{\textbf{Pair}\\\metricup{coverage}} & \shortstack{\textbf{Sel.}\\\metricdown{time}} \\
        \midrule
        \multicolumn{7}{l}{\textbf{\textit{Component ablations}}} \\
        w/o temporal gating & 64.6 & 61.5 & 58.6 & 74.2 & 71.3 & 0.36 \\
        w/o role allocation & 65.1 & 62.3 & 59.1 & 75.0 & 72.8 & 0.37 \\
        w/o core evidence repair & 65.8 & 63.8 & 58.9 & 76.1 & 74.7 & 0.31 \\
        w/o core evidence retention & 66.2 & 64.2 & 60.3 & 76.8 & 75.8 & 0.39 \\
        w/o temporal-pair repair & 66.3 & 64.0 & 60.5 & 77.1 & 75.5 & 0.35 \\
        w/o Constrained Set Refinement & 66.4 & 64.7 & 60.3 & 77.5 & 77.0 & 0.33 \\
        w/o query rewriting & 66.8 & 65.3 & 61.0 & 78.0 & 78.1 & 0.38 \\
        \rowcolor{tessgray}\textbf{TESS (Full)} & \textbf{67.5} & \textbf{66.9} & \textbf{62.2} & \textbf{79.1} & \textbf{79.6} & \textbf{0.41} \\
        \midrule
        \multicolumn{7}{l}{\textbf{\textit{Shared frozen CLIP encoder}}} \\
        Top-$k$ & 61.4 & 57.6 & 54.8 & 68.1 & 58.5 & 0.18 \\
        FOCUS & 62.3 & 59.8 & 56.0 & 72.3 & 65.3 & 0.34 \\
        QCA & 62.8 & 60.6 & 56.3 & 73.5 & 67.2 & 0.38 \\
        \rowcolor{tessgray}\textbf{TESS (Full)} & \textbf{67.5} & \textbf{66.9} & \textbf{62.2} & \textbf{79.1} & \textbf{79.6} & \textbf{0.41} \\
        \bottomrule
    \end{tabularx}
\end{table*}
\noindent\textbf{Evaluation purpose and annotations.}\@
We evaluate evidence preservation on an annotated LongVideoBench subset of 420 questions, including 265 eligible for relation-consistent pair evaluation. Each video--question instance specifies required evidence units, alternative valid intervals, and temporal roles. Matching selected timestamps to these annotations quantifies event and relation coverage independently of downstream answer accuracy.

\noindent\textbf{Semantic coverage.}\@
Evidence recall averages the fraction of required evidence units containing a selected timestamp. Complete coverage measures the proportion of questions for which all required units are represented. Pair coverage measures the proportion of eligible temporal questions whose selected observations cover both reference roles and satisfy the annotated relation. Ordinal and ordering questions additionally require the specified occurrence or sequence. Metric-specific eligibility criteria determine the averaging denominators. Evaluating the complete candidate pool establishes coverage ceilings, providing a reference for the evidence retained by each selector.
\par

\par
\noindent\textbf{Structural diagnostics and controls.}\@
Redundancy measures the mean nearest-neighbor cosine similarity among selected frames, using a fixed evaluation encoder independent of the selection encoder. The maximum gap is the largest interval between consecutive selected timestamps, including video boundaries, normalized by video duration. These measures characterize visual repetition and temporal sparsity. All selectors use identical candidate timestamps and a fixed budget of 64 frames, isolating selection behavior under shared input conditions.

\noindent\textbf{Selection quality.}\@
\Cref{tab:selection-diagnostics} shows that \method achieves the highest evidence recall, complete coverage, and pair coverage, together with the lowest redundancy among the evaluated selectors. Uniform sampling yields a normalized maximum gap of 0.016, complete coverage of 46.4\%, and pair coverage of 41.9\%. The corresponding values for \method are 0.045, 65.0\%, and 79.6\%, indicating more complete semantic and relational evidence despite less uniform temporal spacing. These results support evaluating temporal distribution jointly with annotated event coverage to characterize the evidence available for downstream reasoning.

\par

% Queue Figure S2 on the next page, above the continuing appendix text.
\begin{figure*}[t]
    \centering
    \includegraphics[width=\textwidth]{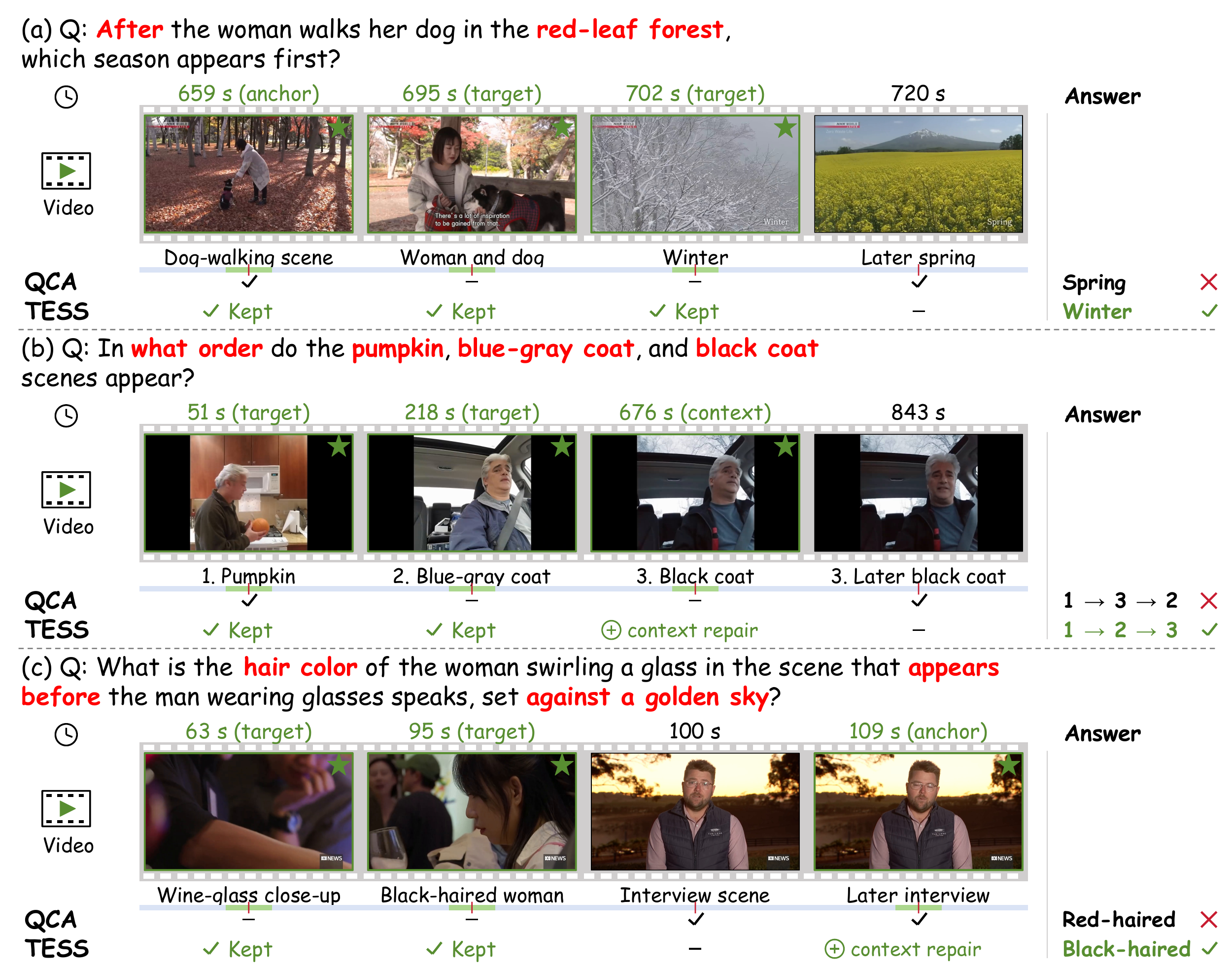}
    % Preserve the comparison figure identifier after moving it earlier.
    \setcounter{figure}{1}
    \caption{\textbf{Relation-conditioned evidence selection in TESS.} Comparisons with QCA under the same 32-frame budget and QA model illustrate (a) subsequent-event identification, (b) event ordering, and (c) preceding-event retrieval. Check marks indicate selected observations, and context-repair annotations identify added evidence. \method answers all three questions correctly.}
    \label{fig:temporal-evidence-cases}
\end{figure*}
\setcounter{figure}{0}

\section{Component and Encoder Controls}
\label{sec:ablation-protocol}
\noindent\textbf{Component definitions.}\@
We use component ablations to examine how relation modeling, evidence allocation, repair, and refinement contribute to the final evidence set. All component ablations retain the same candidate pool and downstream inference settings as the full method. The frame budget and encoders are also held fixed, allowing changes in evidence coverage and QA accuracy to be associated with the omitted operation.

The four principal variants remove temporal gating, role allocation, Core Evidence Repair, or Constrained Set Refinement. Removing temporal gating uses ungated target scores, testing the contribution of relation-conditioned temporal localization. Removing role allocation eliminates role-specific quotas while retaining role scores, testing whether explicit budget allocation preserves complementary observations. Disabling Core Evidence Repair bypasses query, context, and temporal-pair repair, assessing the recovery of evidence omitted during initial selection. Removing Constrained Set Refinement omits the subsequent constrained frame exchanges, assessing the contribution of optimizing the completed evidence set.

Three additional controls distinguish the mechanisms within this pipeline. Disabling core retention permits replacement of every selected frame during repair while preserving the repair budget and scoring criteria, testing the value of retaining initial role representatives. Disabling temporal-pair repair removes pair and neighborhood exchanges while retaining query and context repair, isolating the recovery of compatible event pairs. Disabling query rewriting replaces weighted query responses with the original full-question response throughout selection, repair, and refinement, assessing the contribution of semantic reformulation to evidence retrieval.

\noindent\textbf{Evaluation and timing.}\@
QA performance is evaluated on LongVideoBench with LLaVA-Video-7B and a budget of 64 selected frames. Overall accuracy summarizes benchmark performance, while temporal and long-video subsets examine sensitivity to event relationships and extended temporal context. Temporal annotations identify directional, simultaneous, ordinal, and sequence questions. The annotated diagnostic subset separately measures whether the selected observations retain the required events and their relationships, connecting downstream performance with the composition of the visual evidence.

Selection latency measures the interval from decoded candidates to the final frame set, including feature encoding, query processing, and frame selection. This interval characterizes the selector's computational cost. End-to-end GPU hours additionally account for decoding and downstream QA, providing a complementary measure of the complete inference pipeline.

% Queue the failure illustration alongside its case analysis.

\noindent\textbf{Shared-encoder design.}\@\nopagebreak[4]
To assess selection independently of representation quality, we evaluate Top-$k$, FOCUS, QCA, and \method using identical frozen CLIP features and candidate timestamps. The shared representation makes the construction of the selected frame set the principal difference in this comparison. Jointly examining evidence coverage and QA accuracy then characterizes both the observations retained by each selector and their utility for downstream reasoning.
\par

\begin{figure*}[t]
    \centering
    \includegraphics[width=\textwidth]{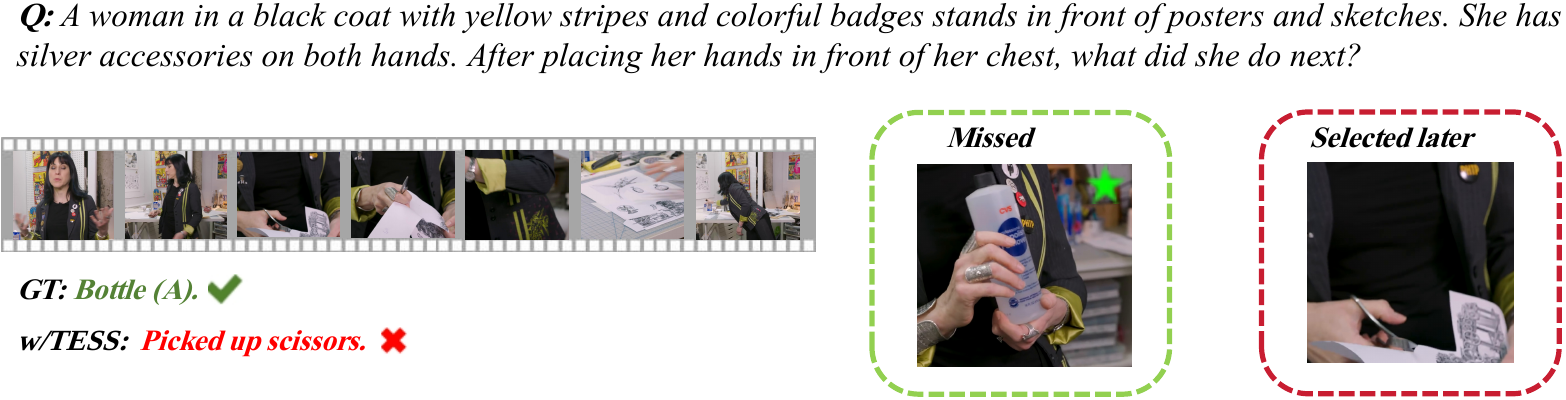}
    \caption{\textbf{Local event omission in an immediate-successor query.} A LongVideoBench example evaluated with LLaVA-Video-7B and 32 selected frames. The strip shows seven selected observations. The green inset shows missed evidence of the white-bottle action, and the red inset shows a selected later scissors action. The omitted bottle action immediately follows the referenced gesture.}
    \label{fig:failure-case}
\end{figure*}
\setcounter{figure}{2}

\noindent\textbf{Component effects.}\@
\Cref{tab:ablation} shows consistent reductions in evidence recall, temporal QA, and pair coverage when individual components are removed. Pair coverage decreases from 79.6\% to 71.3--78.1\% under ablation. Removing temporal gating and role allocation incurs the largest losses of 8.3 and 6.8 points, respectively, supporting the combination of relation-conditioned localization and explicit allocation across evidence roles. Removing Core Evidence Repair and Constrained Set Refinement reduces pair coverage by 4.9 and 2.6 points, indicating complementary benefits from recovering missing observations and refining their joint composition.

The additional controls clarify how these benefits arise. Removing core retention reduces pair coverage, temporal QA, and evidence recall by 3.8, 2.7, and 2.3 points, respectively. Removing temporal-pair repair incurs corresponding losses of 4.1, 2.9, and 2.0 points. Removing query rewriting reduces pair coverage, temporal QA, and evidence recall by 1.5, 1.6, and 1.1 points, respectively. The accompanying decreases in evidence coverage and answer accuracy support preserving role representatives, recovering compatible event pairs, and broadening query-conditioned retrieval as complementary contributions to temporal reasoning.

\noindent\textbf{Shared-encoder results.}\@
Under shared CLIP features, \method achieves 79.6\% pair coverage and exceeds QCA by 12.4, 5.6, and 6.3 points in pair coverage, evidence recall, and temporal QA, respectively. These gains persist when the visual representation is held fixed, supporting the contribution of how frames are selected and organized. Selection takes 0.41 seconds per query, compared with 0.38 for QCA and 0.31--0.39 for component variants. We therefore obtain more complete relational evidence and stronger temporal QA with a modest increase in selection time.

\section{Relation-Conditioned Evidence Selection}
\label{sec:temporal-evidence-cases}
\Cref{fig:temporal-evidence-cases} compares \method and QCA on subsequent-event identification, event ordering, and preceding-event retrieval, highlighting how query-dependent temporal roles determine the evidence required for each question. With the same 32-frame budget and downstream QA model, \method answers all three questions correctly. In case (a), retaining the winter scene after the dog-walking anchor enables identification of the first subsequent season. QCA omits this evidence and predicts the later spring scene. In case (b), \method preserves the intermediate blue-gray-coat scene and supplements a black-coat observation through context repair, retaining the sequence from the pumpkin scene to the blue-gray and black coats. QCA misses the intermediate evidence and predicts an incorrect order. In case (c), \method retains the wine-glass close-up and the black-haired woman before the interview, while context repair adds an interview observation serving as the anchor. These target and anchor observations connect the preceding action to the referenced interview. QCA omits both target observations and incorrectly identifies a red-haired woman.

Across these cases, the required evidence changes with the temporal relation specified by the query. Preserving a subsequent event, an intermediate occurrence, or a preceding action together with its reference context retains the relationships needed for temporal interpretation. These comparisons illustrate how query-dependent evidence organization supports event identification and ordering under the same visual budget.
\par

% Compact the separation below the final single-column figure.
\setlength{\textfloatsep}{10pt plus 2pt minus 2pt}
\section{Evidence Repair and Frame Selection Refinement}
\label{sec:repair-refinement-case}
\begin{figure}[htbp]
    \centering
    \includegraphics[width=\columnwidth]{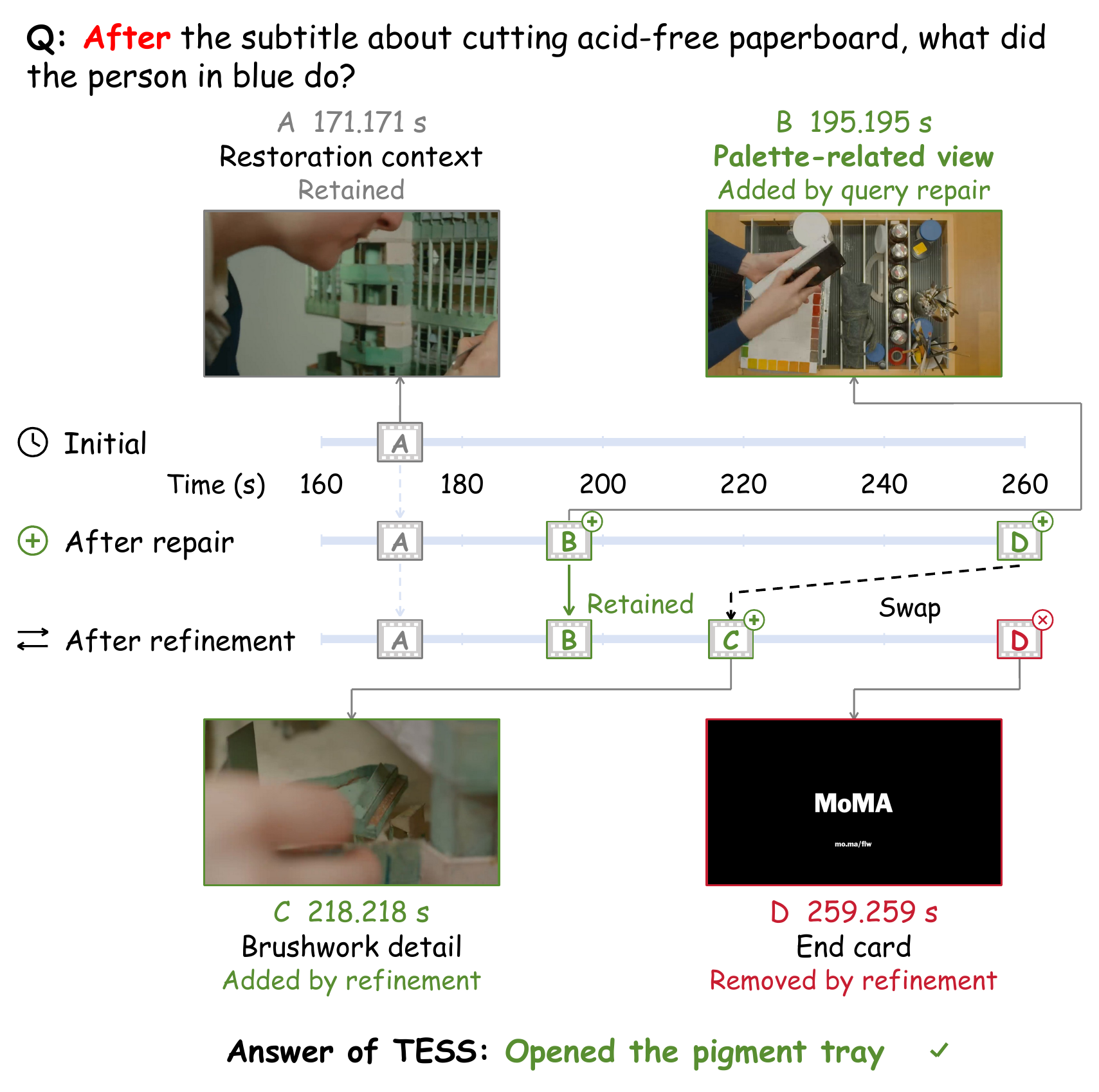}
    \caption{\textbf{Recovery and preservation of target evidence.} The figure illustrates how evidence repair and refinement update the selected frame set. Query repair introduces B while retaining A. Constrained Set Refinement replaces end card D with brushwork frame C while preserving the recovered observation B.}
    \label{fig:repair-refinement-case}
\end{figure}
\Cref{fig:repair-refinement-case} follows a single selection trajectory through Core Evidence Repair and Constrained Set Refinement, focusing on how newly recovered target evidence is retained during later set updates. The three stages share the same candidate pool and a fixed budget of 32 frames. The recorded selection trace identifies the additions, retentions, and replacement involving the four representative observations.

The initial selection preserves the restoration-scene context in frame A at 171.171\,s, while the local pigment-tray view in frame B at 195.195\,s is absent. Query repair supplements the set with B and retains A, adding a targeted observation alongside the existing scene context. This update restores the local visual information associated with the queried action.

Constrained Set Refinement then replaces the end card in frame D at 259.259\,s with the brushwork observation in frame C at 218.218\,s while retaining both A and B. The exchange reallocates a slot from non-action content to an observation of the restoration procedure, preserving the availability of the newly added pigment-tray view. This trajectory illustrates the coordination of the two stages: repair recovers missing target evidence, and subsequent refinement adjusts the composition while retaining that recovery.
\par

\section{Failure Case Analysis}
\label{sec:failure-case}
\Cref{fig:failure-case} illustrates a local evidence omission in an immediate-successor query. The selected sequence skips the white-bottle action immediately following the referenced gesture and retains a later scissors action, leading to an incorrect answer. The selection preserves the person, scene, and overall event order but loses the local adjacency required to identify the next action.

Closely spaced actions with similar visual context challenge precise anchor-boundary localization. In this example, broad temporal coverage preserves the overall sequence while the brief transition relevant to the query remains unrepresented. The failure therefore identifies local event-boundary precision as a complementary requirement to global evidence coverage. We prioritize finer boundary localization and stronger local adjacency constraints during evidence repair to improve the preservation of short events within dense action sequences.
\par

\section{Metrics}
\label{sec:metrics}
\noindent\textbf{QA and evidence coverage.}\@
QA accuracy is the percentage of correctly answered questions, with invalid outputs counted as errors. Overall, Temporal, and Long denote evaluation on the full benchmark, temporal questions, and the long-video subset, respectively. MLVU uses mean accuracy across multiple-choice tasks, giving each task equal weight in the reported aggregate.

Evidence metrics evaluate the selected observations against annotated event intervals and relationships. Evidence recall averages the fraction of required units represented per eligible question. Complete coverage measures the percentage of questions for which every required unit is represented, emphasizing the completeness of the evidence set. Pair coverage measures the percentage of eligible temporal questions whose selected observations represent both reference roles and satisfy the annotated temporal relation. These metrics distinguish partial event retention, complete evidence coverage, and preservation of event relationships.

\noindent\textbf{Structure and efficiency.}\@
Redundancy is the mean nearest-neighbor cosine similarity among selected frames. Maximum gap is the largest timestamp interval, including video boundaries, normalized by video duration. Lower values indicate less visual repetition and smaller uncovered temporal intervals, respectively. These structural measures complement semantic coverage by describing the composition and temporal distribution of the selected set.

The sampled-frame ratio (SFR) measures the proportion of source frames encoded or scored for frame selection. For method $m$, we define
\begin{equation}
    \mathrm{SFR}_m = 100 \times \frac{\sum_j |E_{m,j}|}{\sum_j F_j},
    \label{eq:sfr}
\end{equation}
where $j$ indexes video--question evaluation instances, $F_j$ is the total number of source frames in the video associated with instance $j$, and $E_{m,j}$ is the set of distinct source frames actually encoded or scored by method $m$ for selection. Repeated processing of the same source frame within an instance contributes only once to the numerator. Encoded or scored frames remain included even when they are excluded from the final selected set. Aggregating these counts across instances yields a source-frame-weighted measure of candidate processing, expressed as a percentage.

In the efficiency evaluation, QCA encodes or scores a relatively dense set of candidate frames sampled from each video. \method uniformly samples a smaller candidate pool and extracts CLIP features for every candidate before role allocation and evidence repair within that pool. SFR therefore reflects the candidate sampling configuration of each method, while the fixed 64-frame output budget controls the number of observations supplied to the downstream QA model.

GPU hours aggregate allocated device time across the complete inference pipeline. Selection latency measures feature encoding, query processing, and frame selection after decoding, expressed in seconds per query. Reporting both costs distinguishes selector overhead from the computational demands of complete video question answering.

% Keep the final page in normal reading order: fill the left column first.
\raggedend
\clearpage
\end{document}